\documentclass{article}
\usepackage{ijcai18}

\usepackage{times}
\usepackage{xcolor}
\usepackage{soul}
\usepackage[utf8]{inputenc}
\usepackage[small]{caption}
\usepackage{hyperref}
\usepackage{url}

\usepackage{amsmath}
\usepackage{amssymb}
\usepackage{bm}
\usepackage[ruled,linesnumbered]{algorithm2e}
\usepackage{multirow}
\usepackage{graphicx}

\title{Lifelong Domain Word Embedding via Meta-Learning}

\author{Hu Xu\textnormal{\textsuperscript{1}}, Bing Liu\textnormal{\textsuperscript{1}}, Lei Shu\textnormal{\textsuperscript{1}}\and Philip S. Yu\textnormal{\textsuperscript{1,2}}\\
\textsuperscript{1}{Department of Computer Science, University of Illinois at Chicago, Chicago, IL, USA}\\
\textsuperscript{2}{Institute for Data Science, Tsinghua University, Beijing, China}\\
\{hxu48, liub, lshu3, psyu\}@uic.edu
}

\begin{document}

\maketitle

\begin{abstract}
Learning high-quality domain word embeddings is important for achieving good performance in many NLP tasks. General-purpose embeddings trained on large-scale corpora are often sub-optimal for domain-specific applications. However,  domain-specific tasks often do not have large in-domain corpora for training high-quality domain embeddings. %Although cross-domain embedding methods exist that can leverage general-purpose embeddings to help improve domain-specific embeddings, domain word vectors that have meaning conflicts with the general-purpose embeddings cannot be improved. 
In this paper, we propose a novel \textit{lifelong learning} setting for domain embedding. That is, when performing the new domain embedding, the system has seen many past domains, and it tries to expand the new in-domain corpus by exploiting the corpora from the past domains via meta-learning. The proposed meta-learner characterizes the similarities of the contexts of the same word in many domain corpora, which helps retrieve relevant data from the past domains to expand the new domain corpus. Experimental results show that domain embeddings produced from such a process improve the performance of the downstream tasks. % We also demonstrate that general-purpose embeddings trained from a large corpus are sub-optimal for domain-specific applications.
\end{abstract}

\section{Introduction}
\label{intro}
Learning word embeddings \cite{mnih2007three,turian2010word,mikolov2013efficient,mikolov2013distributed,pennington2014glove}
has received a great deal of attention due to its success in numerous NLP applications, e.g., named entity recognition \cite{sienvcnik2015adapting}, sentiment analysis \cite{maas2011learning} and syntactic parsing \cite{durrett2015neural}.
The key to the success of word embeddings is that a large-scale corpus can be turned into a huge number (e.g., billions) of training examples. 
Two implicit assumptions are often made about the effectiveness of embeddings to down-stream tasks: 
1) the training corpus for embedding is available and much larger than the training data of the down-stream task; 2) the topic (domain) of the embedding corpus is closely aligned with the topic of the down-stream task.
However, many real-life applications do not meet both assumptions.

In most cases, the in-domain corpus is of limited size, which is insufficient for training good embeddings. 
In applications, researchers and practitioners often simply use some general-purpose embeddings trained using a very large general-purpose corpus (which satisfies the first assumption) covering almost all possible topics, e.g., the  GloVe embeddings \cite{pennington2014glove} trained using 840 billion tokens covering almost all topics/domains on the Web. Such embeddings have been shown to work reasonably well in many domain-specific tasks. This is not surprising as the meanings of a word are largely shared across domains and tasks. However, this solution violates the second assumption, which often leads to sub-optimal results for domain-specific tasks, as shown in our experiments.
One obvious explanation for this is that the general-purpose embeddings do provide some useful information for many words in the domain task, but their embedding representations may not be ideal for the domain and in some cases they may even conflict with the meanings of the words in the task domain because words often have multiple senses or meanings. %without in-domain awareness, many out-of-domain representations harmful to in-domain words. 
For example, we have a task in the programming domain, which has the word ``Java''. A large-scale general-purpose corpus, which is very likely to include texts about coffee shops, supermarkets, the Java island of Indonesia, etc., can easily squeeze the room for representing ``Java''' context words like ``function'', ``variable'' or ``Python'' in the programming domain.
This results in a poor representation of the word ``Java'' for the programming task.  

To solve this problem and also the limited in-domain corpus size problem, cross-domain embeddings have been investigated \cite{bollegala-maehara-kawarabayashi:2015:ACL-IJCNLP,yang-lu-zheng:2017:EMNLP2017,bollegala2017think} via transfer learning \cite{pan2010survey}.
These methods allow some in-domain words to leverage the general-purpose embeddings in the hope that the meanings of these words in the general-purpose embeddings do not deviate much from the in-domain meanings of these words. The embeddings of these words can thus be improved. However, these methods cannot improve the embeddings of many other words with domain-specific meanings (e.g., ``Java'').
Further, some words in the general-purpose embeddings may carry meanings that are different from those in the task domain. 

% Thus, the very in-domain information is not expanded.

In this paper, we propose a novel direction for domain embedding learning by expanding the in-domain corpus. The problem in this new direction can be stated as follows:

\textbf{Problem statement}: We assume that the learning system has seen $n$ domain corpora in the past: $D_{1:n}=\{D_1, \dots, D_n\}$, when a new domain corpus $D_{n+1}$ comes with a certain task, the system automatically generates word embeddings for the $(n+1)$-th domain by leveraging some useful information or knowledge from the past $n$ domains.

This problem definition is in the
\textit{lifelong learning} (LL) setting, where the new or $(n+1)$-th task is performed with the help of the knowledge accumulated over the past $n$ tasks \cite{ChenLiu2016}. 
%Silver2013
Clearly, the problem does not have to be defined this way with the domains corpora coming in a sequential manner. It will still work as long as we have $n$ existing domain corpora and we can use them to help with our target domain embedding learning, i.e., the $(n+1)$-th domain.

The main challenges of this problem are 2-fold:
1) how to automatically identify relevant information from the past $n$ domains with no user help, and 2) how to integrate the relevant information into the $(n+1)$-th domain corpus. We propose a meta-learning based system L-DEM (\underline{L}ifelong \underline{D}omain \underline{E}mbedding via \underline{M}eta-learning) to tackle the challenges.

To deal with the first challenge, for a word in the new domain, L-DEM learns to identify similar contexts of the word in the past domains. Here the context of a word means the surrounding words of that word in a domain corpus. We call such context \emph{domain context} (of a word). For this, we introduce a multi-domain meta-learner that can identify similar (or relevant) domain contexts that can be later used in embedding learning in the new domain. To tackle the second challenge, L-DEM augments the new domain corpus with the relevant domain contexts (knowledge) produced by the meta-learner from the past domain corpora and uses the combined data to train the embeddings in the new domain. For example. for word ``Java'' in the programming domain (the new domain), the meta-learner will produce similar domain contexts from some previous domains like programming language, software engineering, operating systems, etc. These domain contexts will be combined with the new domain corpus for ``Java" to train the new domain embeddings. % , and it will not get the contexts from the coffee shop or the Java island of Indonesia domain. 
% , etc. domain but use the knowledge about ``java'' from the OS domain. More importantly, as we see more domains, we gradually \emph{learn} to identify as we see similar domains (like Android domain).

% Besides leveraging the meaning of general words, we humans also learn the specific meaning of a new domain word more smartly. The major difference is that we also have a mechanism to isolate past experience as separated domains.  We accumulate different domain contexts for the same word over many past domains. When a new learning task comes, we quickly identify the new domain contexts and selectively borrow relevant knowledge from past domains.  For example, for the word ``java'' in the programming domain, we never consider the knowledge from the coffee shop domain but use the knowledge about ``java'' from the OS domain. More importantly, as we see more domains, we gradually \emph{learn} to identify as we see similar domains (like Android domain). To shorten the gap with the human learning, we define the following meta-learning \cite{vilalta2002perspective} task of domain embedding, which does not focus on embedding learning of a single domain. 

The main contributions of this paper are as follows.
1) It proposes a novel direction for domain embedding learning, which is a lifelong or continual learning setting and can benefit down-stream learning tasks that require domain-specific embeddings. 
%We are not aware of any existing work on domain embeddings using lifelong learning to accumulate domain knowledge.
2) It proposes a meta-learning approach to leveraging the past corpora from different domains to help generate better domain embeddings. To the best of our knowledge, this is the first meta-learning based approach to helping domain-specific embedding. 3) It experimentally evaluates the effectiveness of the proposed approach. 

\section{Related Works}
\label{rw}
Learning word embeddings has been studied for a long time \cite{mnih2007three}. 
Many earlier methods used complex neural networks \cite{mikolov2013linguistic}. %\cite{collobert2008unified,mikolov2013linguistic}.
More recently, a simple and effective unsupervised model called skip-gram (or word2vec in general) \cite{mikolov2013distributed,mikolov2013linguistic} was proposed to turn a plain text corpus into large-scale training examples without any human annotation.
It uses the current word to predict the surrounding words in a context window. % by maximizing the likelihood of the predictions. 
The learned weights for each word are the embedding of that word.
Although some embeddings trained using large scale corpora are available \cite{pennington2014glove,bojanowski2016enriching}, they are often sub-optimal for domain-specific tasks \cite{bollegala-maehara-kawarabayashi:2015:ACL-IJCNLP,yang-lu-zheng:2017:EMNLP2017,xu:Short,Xu2018pro,xu2017Fun}. 
However, a single domain corpus is often too small for training high-quality embeddings \cite{Xu2018pro}.

Our problem setting is related to \textit{Lifelong Learning} (LL). Much of the work on LL focused on supervised learning \cite{Thrun1996learning,Silver2013,ChenLiu2016}. %ruvolo2013ella,
In recent years, several LL works have also been done for unsupervised learning, e.g., topic modeling \cite{chen2014topic}, information extraction \cite{Mitchell2015} and graph labeling \cite{shu2016lifelong}. 
However, we are not aware of any existing research on using LL for word embedding. Our method is based on meta-learning, which is very different from existing LL methods.
Our work is related to transfer learning and multi-task learning \cite{pan2010survey}. Transfer learning has been used in cross-domain word embeddings \cite{bollegala-maehara-kawarabayashi:2015:ACL-IJCNLP,yang-lu-zheng:2017:EMNLP2017}. However, LL is different from transfer learning or multi-task learning \cite{ChenLiu2016}. 
Transfer learning mainly transfers common word embeddings from general-purpose embeddings to a specific domain. We expand the in-domain corpus with similar past domain contexts identified via meta-learning. 

To expand the in-domain corpus, a good measure of the similarity of domain contexts of the same word from two different domains is needed.
We use meta-learning \cite{thrun2012learning} to learn such similarities.
Recently, meta-learning has been applied to various aspects of machine learning, 
such as learning an optimizer \cite{andrychowicz2016learning}, %learning network configurations \cite{fernando2017pathnet}, 
and learning initial weights for few-shot learning \cite{finn2017model}.
The way we use meta-learning is about domain independent learning \cite{JMLR:v17:15-239}. It learns similarities of domain contexts of the same word. 

\section{Model Overview}
\begin{figure}[t]
    \label{fig:ll}
    \centering    
    \includegraphics[width=3.3in]{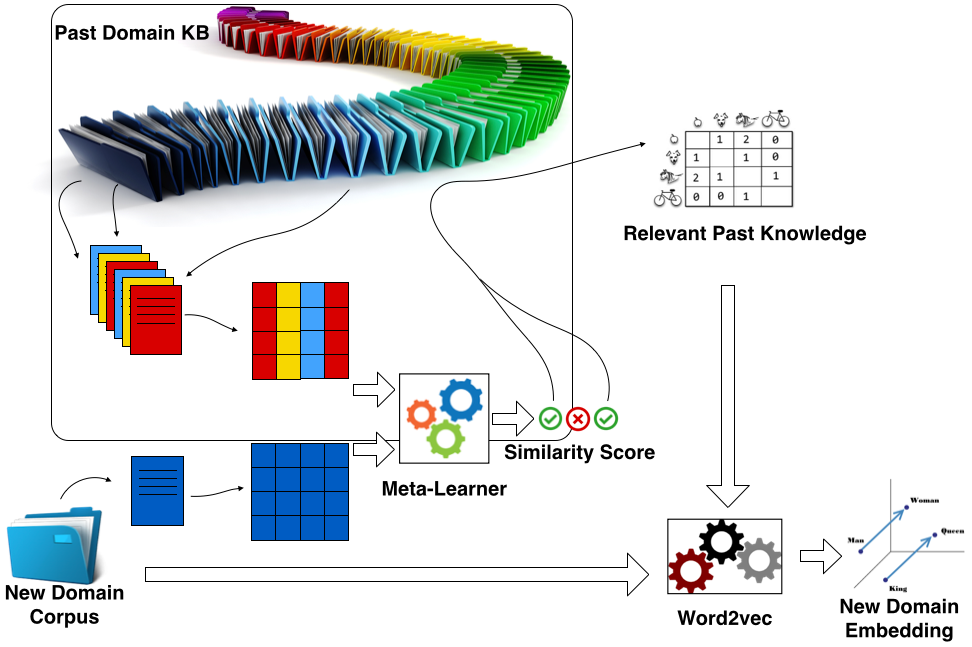}
        \caption{Overview of L-DEM.}
        \label{fig:fr}
    \end{figure}

The proposed L-DEM system is depicted in Figure \ref{fig:fr}.
Given a series of past domain corpora $D_{1:n}=\{D_1, D_2, \dots, D_n\}$, and a new domain corpus $D_{n+1}$, the system learns to generate the new domain embeddings by exploiting the relevant information or knowledge from the past $n$ domains.
Firstly, a base meta-learner $M$ is trained from the first $m$ past domains (not shown in the figure) (see Section 4), which is later used to predict the similarities of \emph{domain contexts} of the same words from two different domains.
Secondly, assuming the system has seen $n-m$ past domain corpora $D_{m+1:n}$, when a new domain $D_{n+1}$ comes, the system produces the embeddings of the $(n+1)$-th domain as follows (discussed in Section 5):
(i) the base meta-learner first is adapted to the $(n+1)$-th domain as $M_{n+1}$ (not shown in the figure) using the $(n+1)$-th domain corpus;
(ii) for each word $w_{i}$ in the new domain, the system uses the adapted meta-learner $M_{n+1}$ to identify every past domain $j$ that has the word $w_{i}$ with domain context similar to $w_{i}$'s domain context in the new domain (we simply call such domain context from a past domain \emph{similar domain context});
(iii) all new domain words' similar domain contexts from all past domain corpora $D_{m+1:n}$ are aggregated. This combined set is called the \textit{relevant past knowledge} and denoted by $\mathcal{A}$;
%Note that this is done for all words in the new domain;
(iv) a modified word2vec model that can take both domain corpus $D_{n+1}$ and the relevant past knowledge of $\mathcal{A}$ is applied to produce the embeddings for the $(n+1)$-th new domain.
Clearly, the meta-learner here plays a central role in identifying relevant knowledge from past domains.
We propose a pairwise model as the meta-learner. 

To enable the above operations, we need a knowledge base (KB), which retains the information or knowledge obtained from the past domains. Once the $(n+1)$-th domain embedding is done, its information is also saved in the KB for future use. We discuss the detailed KB content in Section 5.1. % finds words in the past domains that are similar to the new domain. 
%\begin{figure}[t]
%    \centering    
%    \includegraphics[width=4.5in]{meta_learner.png}
%        \caption{Meta-learning on domain-level word context similarity.}
%        \label{fig:meta}
%    \end{figure}

\section{Base Meta-Learner}
This section describes the base meta-learner, which identifies similar domain contexts. The input to the meta-learner is a pair of word feature vectors (we simply call them \emph{feature vectors}) representing the domain contexts of the same word from two similar / non-similar domains.
The output of the meta-learner is a similarity score of the two feature vectors. %, which can be used to tell whether one pair is more similar than another pair. 
% A desirable meta-learner should be able to handle multiple domains and trained from scratch using the available corpora from some past domains (e.g., 1 to $m$) without the need for manual labeling.

\subsection{Training Examples}
We assume the number of past domains is large and we hold out the first $m$ domains, where $m \ll n$, as the domains to train and test the base meta-learner.
In practice, if $n$ is small, the $m$ domains can be sampled from the $n$ domains.
The $m$ domains are split into 3 disjoint sets: training domains, validation domains, and testing domains.
% The examples from the validation and testing domains are unseen during training. This isolation ensures the base meta-learner to be general.

To enable the meta-learner to predict the similarity score, we need both positive examples (from similar domains) and negative examples (from dissimilar domains).
Since each past domain can be unique (which makes it impossible to have a positive pair from two similar domains), we sub-sample each domain corpus $D_j$ into 2 sub-corpora: $D_{j, k} \sim P(D_i)$, where $1\le j\le m$ and $k=\{1, 2\}$.
This sampling process is done by drawing documents (each domain corpus is a set of documents) uniformly at random from $D_j$.
The number of documents that a domain sub-corpus can have is determined by a pre-defined sub-corpus (file) size (explained in Section 6).
We enforce the same file size across all sub-corpora so feature vectors from different sub-corpora are comparable.

Next, we produce feature vectors from domain sub-corpora.
Given a word $w_{i, j, k}$ (instance of the word $w_{i}$ in the domain sub-corpus $D_{j, k}$), we choose its co-occurrence counts on a fixed vocabulary $V_{\textit{wf}}$ within a context window (similar to word2vec) as the word $w_{i, j, k}$'s feature vector $\mathbf{x}_{w_{i, j, k}}$.
The fixed vocabulary $V_{\textit{wf}}$ (part of the KB used later, denoted as $\mathcal{K}.V_{\textit{wf}}$) is formed from the top-$f$ frequent words over $m$ domain corpora.
This is inspired by the fact that an easy-to-read dictionary (e.g., Longman dictionary) uses only a few thousand words to explain all words of a language.
A pair of feature vectors $(\mathbf{x}_{w_{i, j, k}}, \mathbf{x}_{w_{i, j, k'}})$ with $k \neq k'$, forms a postive example; 
whereas $(\mathbf{x}_{w_{i, j, k}}, \mathbf{x}_{w_{i, j', k}})$ with $j\neq j'$ forms a negative example.
Details of settings are in Section 6.

\subsection{Pairwise Model of the Meta-learner}
We train a small but efficient pairwise model (meta-learner) to learn similarity score. 
Making the model small but high-throughput is crucial.
This is because the meta-learner is required in a high-throughput inference setting, where every word from a new domain needs to have context similarities with the same word from all past domains. 

The proposed pairwise model has only four layers. 
One shared fully-connected layer (with $l_1$-norm) is used to learn two continuous representations from two (discrete) input feature vectors.
A matching function is used to compute the representation of distance in a high-dimentional space.
Lastly, a fully-connected layer and a sigmoid layer are used to produce the similarity score.
The model is parameterized as follows:
\begin{equation}
\sigma \big( \bm{W}_2 \cdot \text{abs}\big( ( \bm{W}_1 \cdot \frac{\mathbf{x}_{w_{i, j, k}}}{ |\mathbf{x}_{w_{i, j, k}}|_1 }) - (\bm{W}_1 \cdot \frac{\mathbf{x}_{w_{i, j', k'}} }{ |\mathbf{x}_{w_{i, j', k'}}|_1 } ) \big) + b_2 \big) ,
\end{equation}
where $|\cdot|_1$ is the $l_1$-norm, $\text{abs}(\cdot)$ computes the absolute value of element-wise subtraction ($-$) as the matching function, $\bm{W}$s and $b$ are weights and $\sigma (\cdot)$ is the sigmoid function.
The majority of trainable weights resides in $\bm{W}_1$, which learns continuous features from the set of $f$ context words.
These weights can also be interpreted as a general embedding matrix over $V_{\textit{wf}}$. 
These embeddings (not related to the final domain embeddings in Section \ref{sec:aet}) help to learn the representation of domain-specific words.
As mentioned earlier, we train the base meta-learner $M$ over a hold-out set of $m$ domains.
We further fine-tune the base meta-learner using the new domain corpus for its domain use, as described in the next section.

\section{Embedding Using Past Relevant Knowledge}
We now describe how to leverage the base meta-learner $M$, the rest $n-m$ past domain corpora, and the new domain corpus $D_{n+1}$ to produce the new domain embeddings.

\subsection{Identifying Context Words from the Past}
When it comes to borrowing relevant knowledge from past domains, the first problem is what to borrow.
It is well-known that the embedding vector quality for a given word is determined by the quality and richness of that word's contexts.
We call a word in a domain context of a given word a \emph{context word}.
So for each word in the new domain corpus, we should borrow all context words from that word's similar domain contexts.
The algorithm for borrowing knowledge is described in Algorithm \ref{alg:ll}, which finds relevant past knowledge $\mathcal{A}$ (see below) based on the
knowledge base (KB) $\mathcal{K}$ and the new domain corpus $D_{n+1}$.

The KB $\mathcal{K}$ has the following pieces of information:
(1) the vocabulary of top-$f$ frequent words $\mathcal{K}.V_{\textit{wf}}$ (as discussed in Section 4.1), 
(2) the base meta-learner $\mathcal{K}.M$ (discussed in Section 4.2),
and (3) domain knowledge $\mathcal{K}_{m+1:n}$.
The domain knowledge has the following information:
(i) the vocabularies $V_{m+1:n}$ of past $n-m$ domains,
(ii) the sets of past word domain contexts $C_{m+1:n}$ from the past $n-m$ domains, where each $C_j$ is a set of key-value pairs $(w_{i,j}, \mathcal{C}_{w_{i,j} } )$ and $\mathcal{C}_{w_{i,j} } $ is a list of context words\footnote{We use list to simplify the explanation. In practice, bag-of-word representation should be used to save space.} for word $w_i$ in the $j$-th domain, 
and (iii) the sets of feature vectors $E_{m+1:n}$ of past $n-m$ domains, where each set $E_{j}=\{ \mathbf{x}_{w_{i, j, k}} | w_i \in V_{j} \text{ and } k=\{1, 2\} \}$.

The relevant past knowledge $\mathcal{A}$ of the new domain is the aggregation of all key-value pairs $(w_t, \mathcal{C}_{w_t})$, where $\mathcal{C}_{w_t}$ contains all similar domain contexts for $w_t$. % except that $\mathcal{A}$ merges all context words of the same word $w_t$ from different past domains together.

Algorithm \ref{alg:ll} retrieves the past domain knowledge in line 1.
Lines 2-4 prepare the new domain knowledge.
The BuildFeatureVector function produces a set of feature vectors as $E_{n+1}=\{ \mathbf{x}_{w_{i, n+1, k}} | w_i \in V_{j} \text{ and } k=\{1, 2\}\}$ over two sub-corpora of the new domain corpus $D_{n+1}$.
The ScanContextWord function builds a set of key-value pairs, where the key is a word from the new domain $w_{i, n+1}$ and the value $\mathcal{C}_{w_{i,n+1} } $ is a list of context words for the word $w_{i, n+1}$ from the new domain corpus. We use the same size of context window as the word2vec model.

\subsubsection{Adapting Meta-learner}
In line 5, AdaptMeta-learner adapts or fine-tunes the base meta-learner $\mathcal{K}.M$ to produce an adapted meta-learner $M_{n+1}$ for the new domain.
A positive tuning example is sampled from two sub-corpora of the new domain $(\mathbf{x}_{w_{i, n+1, 1}}, \mathbf{x}_{w_{i, n+1, 2}})$ in the same way as described in Section 4.1.
A negative example is exampled as $(\mathbf{x}_{w_{i, n+1, 1}}, \mathbf{x}_{w_{i, j, 2}})$, where $m+1 \le j \le n$.
%So one from the new domain and one from one of the past domain.
%All examples are split into a training set, validation set, and testing set. 
The initial weights of $M_{n+1}$ are set as the trained weights of the base meta-learner $M$. 
% We evaluate the adapted meta-learner in Section 6. 
\begin{algorithm}[t]
\LinesNumbered
\DontPrintSemicolon
\caption{Identifying Context Words from the Past}
\label{alg:ll}
\SetKwInOut{Input}{Input} 
\SetKwInOut{Output}{Output} 
\SetKwRepeat{Do}{do}{while}
\Input{a knowledge base $\mathcal{K}$ containing a vocabulary $\mathcal{K}.V_{\textit{wf}}$, a base meta-learner $\mathcal{K}.M$, and domain knowledge $\mathcal{K}_{m+1:n}$; \\a new domain corpus $D_{n+1}$.}
\Output{relevant past knowledge $\mathcal{A}$, where each element is a key-value pair $(w_t, \mathcal{C}_{w_t})$ and $\mathcal{C}_{w_t}$ is a list of context words from all similar domain contexts for $w_t$. }
\BlankLine
\BlankLine
$(V_{m+1:n}, C_{m+1:n}, E_{m+1:n}) \gets \mathcal{K}_{m+1:n}$ \;
$V_{n+1} \gets \text{BuildVocab}(D_{n+1})$ \;
$C_{n+1} \gets \text{ScanContextWord}(D_{n+1}, V_{n+1})$ \;
$E_{n+1} \gets \text{BuildFeatureVector}(D_{n+1}, \mathcal{K}.V_{\textit{wf}})$ \;
$M_{n+1} \gets \text{AdaptMeta-learner}(\mathcal{K}.M, E_{m+1:n}, E_{n+1})$ \;
$\mathcal{A} \gets \emptyset$ \;
%$W \gets \{\emptyset\}$\;
\For{$(V_j, C_j, E_j) \in (V_{m+1:n}, C_{m+1:n}, E_{m+1:n})$}{
    $O \gets V_j \cap V_{n+1}$  \;
    $F \gets \big\{(\mathbf{x}_{o, j, 1}, \mathbf{x}_{o, n+1, 1})|$  $\text{ }\text{ }\text{ }\text{ }\text{ }\text{ }\text{ }\text{ }\text{ }\text{ } o \in O \text{ and } \mathbf{x}_{o, j, 1} \in E_j \text{ and } \mathbf{x}_{o, n+1, 1} \in E_{n+1} \big\}$ \;
    %$F_j \gets \text{RetrieveFeatureVector}(E_j, O)$ \;
    %$F_{n+1} \gets \text{RetrieveFeatureVector}(E_{n+1}, O)$ \;
    $S \gets M_{n+1}.\text{inference}( F )$ \;
    $O \gets \{o| o\in O \text{ and } S[o]\ge \delta \}$ \;
    \For{$o \in O$}{
    %    \If{$S[o] \ge \delta $}{
        $\mathcal{A}[o].\text{append}(C_j[o] )$ \;
    %    }
    }
    %$ \mathcal{A} \gets \text{MergeCo-occurrence}(\mathcal{A}, P)$ \;
}
%$W \gets \text{InvertbyDomainIndex}(W)$ \;

%\For{$(D_i, O_i) \in (D_{m:n}, W_{m:n})$}{
%    $ \mathcal{A} \gets \mathcal{A} \cup \text{ScanCo-Occurrence}(D_i, O_i)$ \;
%}
$\mathcal{K}_{n+1} \gets (V_{n+1}, C_{n+1}, E_{n+1}) $ \;
\Return{ $\mathcal{A}$}
\end{algorithm}

\subsubsection{Retriving Relevant Past Knowledge}
Algorithm \ref{alg:ll} further produces the relevant past knowledge $\mathcal{A}$ from line 6 through line 16.
Line 6 defines the variable that stores the relevant past knowledge.
Lines 7-15 produce the relevant past knowledge $\mathcal{A}$ from past domains.
The For block handles each past domain sequentially.
Line 8 computes the shared vocabulary $O$ between the new domain and the $j$-th past domain.
After retrieving the sets of feature vectors from the two domains in line 9, the adapted meta-learner uses its inference function (or model) to compute the similarity scores on pairs of feature vectors representing the same word from two domains (line 10).
The inference function can parallelize the computing of similarity scores in a high-throughput setting (e.g., GPU inference) to speed up.
Then we only keep the words from past domains with a score higher than a threshold $\delta$ at line 11.
%This threshold controls the quality of the words with similar contexts $O$.
Lines 12-14 aggregate the context words for each word in $O$ from past word domain contexts $C_j$.
Line 16 simply stores the new domain knowledge for future use.
Lastly, all relevant past knowledge $\mathcal{A}$ is returned.
%Further, the computed co-occurrence can also be cached into knowledge base $\mathcal{K}$ to speed up the further retrieving process.  

\subsection{Augmented Embedding Training}
\label{sec:aet}
We now produce the new domain embeddings via a modified version of the skip-gram model \cite{mikolov2013distributed} that can take both the new domain corpus $D_{n+1}$ and the relevant past knowledge $\mathcal{A}$.
Given a new domain corpus $D_{n+1}$ with the vocabulary $V_{n+1}$, the goal of the skip-gram model is to learn a vector representation for each word $w_{i} \in V_{n+1}$ in that domain
(we omit the subscript $_{n+1}$ in $w_{i, n+1}$ for simplicity).
Assume the domain corpus is represented as a sequence of words $D_{n+1}=(w_1, \dots, w_T)$, the objective of the skip-gram model maximizes the following log-likelihood:
\begin{equation}
\label{eq:sg}
\begin{split}
\mathcal{L}_{D_{n+1}} =\sum_{t=1}^{T} \big( \sum_{w_c \in \mathcal{W}_{w_{t}} } \big(\log \sigma (\bm{u}_{w_t}^T\cdot \bm{v}_{w_c}) \\
+ \sum_{w_{c'} \in \mathcal{N}_{w_t} } \log \sigma(-\bm{u}_{w_t}^T\cdot \bm{v}_{w_{c'}} ) \big) \big) , 
\end{split}
\end{equation}
where $\mathcal{W}_{w_t}$ is the set of words surrounding word $w_t$ in a fixed context window;
$\mathcal{N}_t$ is a set of words (negative samples) drawn from the vocabulary $V_{n+1}$ for the $t$-th word;
$\bm{u}$ and $\bm{v}$ are word vectors (or embeddings) we are trying to learn.
The objective of skip-gram on data of relevant past knowledge $\mathcal{A}$ is as follows:
\begin{equation}
\begin{split}
\mathcal{L}_{\mathcal{A}}=\sum_{(w_t, \mathcal{C}_{w_t} ) \in \mathcal{A}} \big( \sum_{w_c \in \mathcal{C}_{w_t}} \big( \log \sigma (\bm{u}_{w_t}^T\cdot \bm{v}_{w_c}) \\
+ \sum_{w_{c'} \in \mathcal{N}_{w_t} } \log \sigma(-\bm{u}_{w_t}^T\cdot \bm{v}_{w_{c'}} ) \big) \big).
\end{split}
\end{equation}
Finally, we combine the above two objective functions:
\begin{equation}
\begin{split}
\mathcal{L}'_{D_{n+1}}=\mathcal{L}_{D_{n+1}} + \mathcal{L}_{\mathcal{A}}.
\end{split}
\end{equation}
We use the default hyperparameters of skip-gram model \cite{mikolov2013distributed} to train the domain embeddings.

\section{Experimental Evaluation}
\label{sec:exp}
%We now evaluate the effectiveness of our DEM approach. 
Following \cite{nayak2016evaluating}, %,faruqui2016problems}, 
we use the performances of down-stream tasks to evaluate the proposed method. 
We do not evaluate the learned embeddings directly as in \cite{mikolov2013distributed,pennington2014glove} because domain-specific dictionaries of similar / non-similar words are generally not available. Our down-stream tasks are text classification that usually requires fine-grained domain embeddings.

\subsection{Datasets}
We use the Amazon Review datasets from \cite{HeMcA16a}, which is a collection of multiple-domain corpora. We consider each second-level category (the first level is department) as a domain and aggregate all reviews under each category as one domain corpus. This ends up with a rather diverse domain collection. 
We first randomly select 56 ($m$) domains as the first $m$ past domains to train and evaluate the base meta-learner.
Then from rest domains, we sample three random collections with 50, 100 and 200 ($n-m$) domains corpora, respectively, as three settings of past domains.
These collections are used to test the performance of different numbers of past domains.
Due to the limited computing resource, we limit each past domain corpus up to 60 MB.
%Then we deliberately pick the ``Laptops'' domain as the new domain for aspect extraction task since the annotation is on Laptop reviews.
We further randomly selected 3 rest domains (\emph{Computer Components} (CC), \emph{Kitchen Storage and Organization} (KSO) and \emph{Cats Supply} (CS)) as new domains for down-stream tasks. These give us three text classification problems, which have 13, 17, and 11 classes respectively. The tasks are topic-based classification rather than sentiment classification. 
Since the past domains have different sizes (many have much less than 60 MB) and many real-world applications do not have big in-domain corpora, we set the size of the new domain corpora to be 10 MB and 30 MB to test the performance in the two settings.
%(?? the past domain corpora are much larger which is inconsistent with the target domain size for lifelong learning. They should be roughly the same size).
%The second dataset is the Stack Exchange dataset \footnote{\url{https://stackexchange.com/} }. 
%It contains a large scale community QAs, including Stack Overflow (a popular community QA for programmers) and many other diverse communities, such as maths, physicals, psychology, writing, etc.
%We consider each community as a domain.
%This ends up with 88 domains in total. 
\subsection{Evaluation of Meta-Learner}
\begin{table}[t]
\begin{center}

\begin{tabular}{l || c | c | c }
\hline
 & CC & KSO & CS\\
\hline
\hline
10MB & 0.832 & 0.841 & 0.856 \\
30MB & 0.847 & 0.859 & 0.876 \\
\hline

\end{tabular}
\caption{F1-score of positive predictions of the adapted meta-learner on 3 new domains: Computer Components (CC), Kitchen Storage and Organization (KSO) and Cats Supply (CS).}
\end{center}
\label{table:fine-tune}
\end{table}

We select the top $f=5000$ words from all 56 domains' corpora as word features.
Then we split the 56 domains into 39 domains for training, 5 domains for validation and 12 domains for testing.
So the validation and testing domain corpora have no overlap with the training domain corpora.
%This leads to a more general base meta-learner for many unseen new domains.
We sample 2 sub-corpora for each domain and limit the size of each sub-corpus to 10 MB. We randomly select 2000, 500, 1000 words from each training domain, validation domain, and testing domain, respectively, and ignore words with all-zero feature vectors to obtain pairwise examples. 
The testing 1000 words are randomly drawn and they have 30 overlapping words with the training 2000 words, but not from the same domains. So in most cases, it's testing the unseen words in unseen domains.
We set the size of a context window to be 5 when building feature vectors.
This ends up with 80484 training examples, 6234 validation examples, and 20740 test examples.
For comparison, we train a SVM model as a baseline.
The F1-score (for positive pairs) of SVM is 0.70, but the F1-score of the proposed base meta-learner model is \textbf{0.81}.

To adapt the base meta-learner for each new domain. We sample 3000 words from each new domain, which results in slightly fewer than 6000 examples after ignoring all-zero feature vectors.
We select 3500 examples for training, 500 examples for validation and 2000 examples for testing.
The F1-scores on the test data is shown in Table 1.
Finally, we empirically set $\delta=0.7$ as the threshold on the similarity score in Algorithm 1, which roughly doubled the number of training examples from the new domain corpus. 
The size of the context window for building domain context is set to 5, which is the same as word2vec.
% CAN WE SAY SMALL THRESHOLD ENDS WITH BAD PERFORMANCE LARGER THRESHOLD ENDS WITH SMALL RELEVANT KNOWLEDGE ?

\begin{table}[t]
\begin{center}

\scalebox{0.9}{

\begin{tabular}{l || c | c | c }
\hline
 & CC(13) & KSO(17) & CS(11)\\
\hline
\hline
NE  &  0.596  &  0.653  &  0.696 \\
fastText  &  0.705  &  0.717  &  0.809 \\
GoogleNews  &  0.76  &  0.722  &  0.814 \\
GloVe.Twitter.27B  &  0.696  &  0.707  &  0.80 \\
GloVe.6B  &  0.701  &  0.725  &  0.823 \\
GloVe.840B  &  0.803  &  0.758  &  0.855 \\
ND 10M  &  0.77  &  0.749  &  0.85 \\
ND 30M  &  0.794  &  0.766  &  0.87 \\
200D + ND 30M  &  0.795  &  0.765  &  0.859 \\
\hline
L-DENP 200D + ND 30M & 0.806 & 0.762 & 0.870 \\
\hline
L-DEM 200D + ND 10M  &  0.791  &  0.761  &  0.872 \\
L-DEM 50D + ND 30M  &  0.795  &  0.768  &  0.868 \\
L-DEM 100D + ND 30M  &  0.803  &  0.773  &  0.874 \\
L-DEM 200D + ND 30M  &  \textbf{0.809}  &  \textbf{0.775}  &  \textbf{0.883} \\
\hline

\end{tabular}
}
\caption{Accuracy of different embeddings on classification tasks for 3 new domains (numbers in parenthesis: the number of classes)}
\end{center}
\label{table:pc}
\end{table}

\subsection{Baselines and Our System}
% We evaluate the following baseline embeddings.\\
Unless explicitly mentioned, the following embeddings have 300 dimensions, which are the same size as many pre-trained embeddings (GloVec.840B \cite{pennington2014glove} or fastText English Wiki\cite{bojanowski2016enriching}).\\
\textbf{No Embedding (NE)}: This baseline does not have any pre-trained word embeddings. The system randomly initializes the word vectors and train the word embedding layer during the training process of the down-stream task. % Note that only in this baseline do we allow embeddings trainable.
\\
\textbf{fastText}: This baseline uses the lower-cased embeddings pre-trained from English Wikipedia using fastText \cite{bojanowski2016enriching}. We lower the cases of all corpora of down-stream tasks to match the words in this embedding. \\
% Note that although the corpus of Wikipedia contains a wide spectrum of domains covering almost everything of human knowledge, the size of a corpus for a specific domain (e.g, a product) may not be large enough. The total amount of Wikipedia is just several billions of tokens, which is on the same scale as Amazon Review datasets (8 billion tokens).\\
\textbf{GoogleNews}: This baseline uses the pre-trained embeddings from word2vec \footnote{https://code.google.com/archive/p/word2vec/} based on part of the Google News dataset, which contains 100 billion words.\\
\textbf{GloVe.Twitter.27B}: This embedding set is pre-trained using GloVe\footnote{https://nlp.stanford.edu/projects/glove/} based on Tweets of 27 billion words. This embedding is lower-cased and has 200 dimensions.\\
\textbf{GloVe.6B}: This is the lower-cased embeddings pre-trained from Wikipedia and Gigaword 5, which has 6 billion tokens. \\
\textbf{GloVe.840B}: This is the cased embeddings pre-trained from Common Crawl corpus, which has 840 billion tokens. 
%This embedding corpus is the largest one among all embeddings.
This corpus contains almost all web pages available before 2015. 
We show that the embeddings produced from this very general corpus are sub-optimal for our domain-specific tasks.\\
\textbf{New Domain 10M (ND 10M)}: This is a baseline embedding pre-trained only from the new domain 10 MB corpus. 
We show that the embeddings trained from a small corpus alone are not good enough.\\
\textbf{New Domain 30M (ND 30M)}: This is a baseline embedding pre-trained only from the new domain 30 MB corpus. We increase the size of the new domain corpus to 30 MB to see the effect of the corpus size. \\
\textbf{200 Domains + New Domain 30M (200D + ND 30M)}: The embedding set trained by combining the corpora from all past 200 domains and the new domain. We use this baseline to show that using all past domain corpora may reduce the performance of the down-stream tasks. \\
%and how to smartly select relevant word contexts from past domains is crucial.\\
\textbf{L-DENP 200D + ND 30M}: This is a \underline{N}on-\underline{P}arametric variant of the proposed method. We use TFIDF as the representation for a sentence in past domains and use cosine as a non-parametic function to compute the similarity with the TFIDF vector built from the new domain corpus. 
We report the results on a similarity threshold of 0.18, which is the best threshold ranging from 0.15 to 0.20.\\
\textbf{L-DEM Past Domains + New Domain (L-DEM [P]D + ND [X]M)}: These are different variations of our proposed method L-DEM. For example, ``L-DEM 200D + ND 30M'' denotes the embeddings trained from a 30MB new domain corpus and the relevant past knowledge from 200 past domains.

\subsection{Down-stream Tasks and Experiment Results}
As indicated earlier, we use classification tasks from 3 new domains (``Computer Components'', ``Cats Supply'' and ``Kitchen Storage and Organization'') to evaluate the embeddings produced by our system and compare them with those of baselines. 
These 3 new domains have 13, 17 and 11 classes (or product types), respectively.
For each task, we randomly draw 1500 reviews from each class to make up the experiment data, from which we keep 10000 reviews for testing (to make the result more accurate) and split the rest 7:1 for training and validation, respectively.
All tasks are evaluated on accuracy.
We train and evaluate each task on each system 10 times (with different initializations) and average the results.

For each task, we use an embedding layer to store the pre-trained embeddings.
We freeze the embedding layer during training, so the result is less affected by the rest of the model and the training data.
To make the performance of all tasks consistent, 
we apply the same Bi-LSTM model \cite{hochreiter1997long} on top of the embedding layer to learn task-specific features from different embeddings.
The input size of Bi-LSTM is the same as the embedding layer and the output size is 128.
All tasks use many-to-one Bi-LSTMs for classification purposes.
In the end, a fully-connected layer and a softmax layer are applied after Bi-LSTM, with the output size specific to the number of classes of each task.
We apply dropout rate of 0.5 on all layers except the last one and use Adam \cite{kingma2014adam} as the optimizer.

Table 2 shows the main results. 
%, we can see that the performance of different classification tasks varies a lot. This is mostly caused by the number of classes in each task.
We observe that the proposed method L-DEM 200D + ND 30M performs the best.
The difference in the numbers of past domains indicates more past domains give better results.
The GloVe.840B trained on 840 billion tokens does not perform as well as embeddings produced by our method. 
%(?? add a break down analysis and pave the way for DE-TC) 
GloVe.840B's performance on the CC domain is close to our method indicating mixed-domain embeddings for this domain is not bad and this domain is more general.  
% But it performs very well on Computer Components, which means this domain is relatively general.
Combining all past domain corpora together with the new domain corpus (200D + ND 30M) makes the result worse than not using the past domains at all (ND 30M).
This is because the diverse 200 domains are not similar to the new domains.
The L-DENP 200D + ND 30M performs poorly indicating the proposed parametric meta-learner is useful, except the CC domain which is more general.

\subsection{Additional Experiments}
\begin{table}[t]
\begin{center}
\scalebox{0.80}{
\begin{tabular}{l || c | c | c }
\hline
 & CC(13) & KSO(17) & CS(11)\\
\hline
\hline
GloVe.840B\&ND 30M  &  0.811  &  0.78  &  0.885 \\
GloVe.840B\&L-DEM 200D+30M  &  \textbf{0.817}  &  \textbf{0.783}  &  \textbf{0.887} \\
\hline
\end{tabular}
}
\end{center}
\label{table:concat}
\caption{Results of concatenated embeddings with GloVe.840B}
\end{table}

Note that we did not compare with the existing transfer learning methods~\cite{bollegala2017think,bollegala-maehara-kawarabayashi:2015:ACL-IJCNLP,yang-lu-zheng:2017:EMNLP2017} as our approaches focus on domain-specific words in a lifelong learning setting, which do not need the user to provide the source domain(s) that are known to be similar to the target domain. One approach to leveraging existing embeddings is to concatenate pre-trained embeddings with domain-specific embeddings\footnote{Note the ideal LL setting is to perform L-DEM over all domain corpora of the pre-trained embeddings.}. To demonstrate our method further improves the domain-specific parts of the down-stream tasks, we evaluate two methods: (1) GloVe.840B\&ND 30M, which concatenates new domain only embeddings with GloVe.840B; (2) GloVe.840B\&L-DEM 200D + ND 30M, which concatenates our proposed embeddings with GloVe.840B. 
As shown in Table 3, concatenating embeddings improve the performance.
% (?? I removed this review purpose footnote)
%\footnote{\footnotesize{Note that it is not appropriate to compare these results with those in Table 2 as our method uses much less data than GloVe.840B.}}. 
Our method boosts the domain-specific parts of the embeddings further.
% While existing cross-domain embedding methods can only use the 30 MB corpus of the new domain, our method allows those methods to leverage relevant knowledge from past domains. 

\section{Conclusions}
In this paper, we formulated a domain word embedding learning process. 
Given many previous domains and a new domain corpus, the proposed method can generate new domain embeddings by leveraging the knowledge in the past domain corpora via a meta-learner. 
%The meta-learner basically identify similar domain contexts from the past domains to augment the new domain corpus, which enables the system to produce better embeddings in the new domain. 
Experimental results show that our method is highly promising. 
%\subsubsection*{Acknowledgments}


\section*{Acknowledgments}
This work is supported in part by NSF through grants IIS-1526499, IIS-1763325, IIS1407927, CNS-1626432 and NSFC 61672313, and a gift from Huawei Technologies. 
% We also thank anonymous reviewers for their valuable feedbacks.

\bibliographystyle{named}
\bibliography{ijcai18}

\end{document}